\pdfoutput=1

\documentclass[10pt,twocolumn,letterpaper]{article}

\usepackage{cvpr}              

\usepackage{multirow} 
\usepackage{booktabs} 
\usepackage{diagbox} 
\usepackage{makecell} 
\usepackage{pifont}   

\definecolor{cvprblue}{rgb}{0.21,0.49,0.74}
\usepackage[pagebackref,breaklinks,colorlinks,allcolors=cvprblue]{hyperref}

\def\paperID{207} 
\def\confName{CVPR}
\def\confYear{2025}

\title{Fast and Accurate Gigapixel Pathological Image Classification \\ with Hierarchical Distillation Multi-Instance Learning}

\author{
Jiuyang Dong$^{1}$, \and \hspace{-1cm} 
Junjun Jiang$^{1}$\thanks{Corresponding author: Junjun Jiang, Yongbing Zhang}, \and \hspace{-1cm} 
Kui Jiang$^{1}$, \and \hspace{-1cm} 
Jiahan Li$^{1}$ ,\and \hspace{-1cm}
Yongbing Zhang$^{2}$\footnotemark[1] \\
$\textsuperscript{1}$Harbin Institute of Technology, 
$\textsuperscript{2}$Harbin Institute of Technology, Shenzhen\\ 
{\tt\small \{jiuyang.dong, jiahan.li\}@stu.hit.edu.cn,}\\
{\tt\small \{jiangjunjun, jiangkui, ybzhang08\}@hit.edu.cn}\\
}

\begin{document}
\maketitle
\begin{abstract}
Although multi-instance learning (MIL) has succeeded in pathological image classification, it faces the challenge of high inference costs due to processing numerous patches from gigapixel whole slide images (WSIs).
To address this, we propose HDMIL, a hierarchical distillation multi-instance learning framework that achieves fast and accurate classification by eliminating irrelevant patches.
HDMIL consists of two key components: the dynamic multi-instance network (DMIN) and the lightweight instance pre-screening network (LIPN). DMIN operates on high-resolution WSIs, while LIPN operates on the corresponding low-resolution counterparts.
During training, DMIN are trained for WSI classification while generating attention-score-based masks that indicate irrelevant patches.
These masks then guide the training of LIPN to predict the relevance of each low-resolution patch.
During testing, LIPN first determines the useful regions within low-resolution WSIs, which indirectly enables us to eliminate irrelevant regions in high-resolution WSIs, thereby reducing inference time without causing performance degradation.
In addition, we further design the first Chebyshev-polynomials-based Kolmogorov-Arnold classifier in computational pathology, which enhances the performance of HDMIL through learnable activation layers.
Extensive experiments on three public datasets demonstrate that HDMIL outperforms 
previous state-of-the-art methods, e.g., achieving improvements of 3.13\% in AUC while reducing inference time by 28.6\% on the Camelyon16 dataset.
The project is available at https://github.com/JiuyangDong/HDMIL.
\end{abstract}

\section{Introduction}
\label{sec:intro}
\begin{figure}[!h]
\centering
\includegraphics[width=0.48\textwidth]{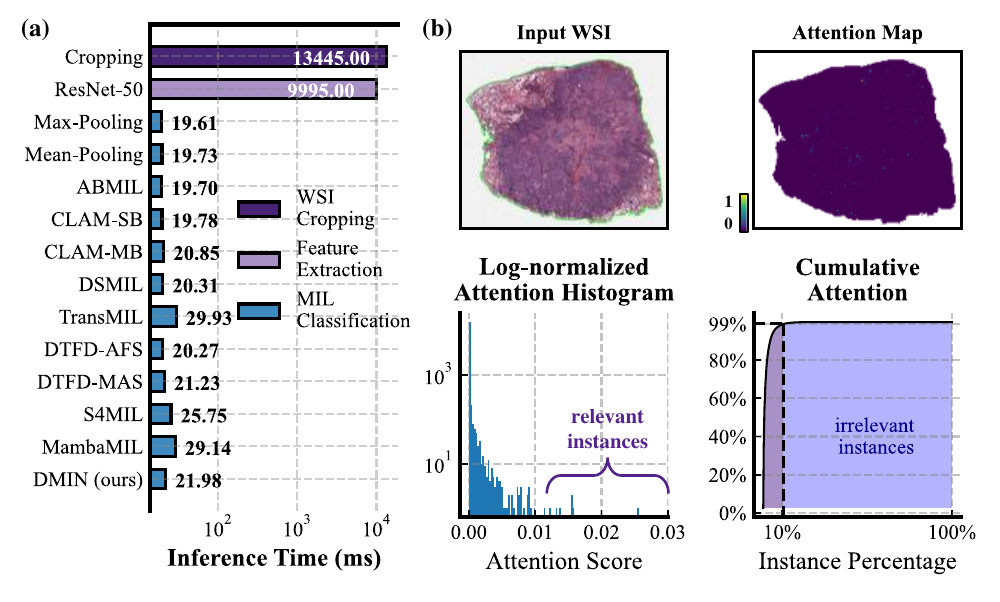}
\caption{
\textbf{What makes inference slow?}
(a) \textbf{Time-consuming data pre-processing}: 
After comparing the time required for data pre-processing (WSI cropping, feature extraction) and MIL network classification, it is clear that data pre-processing is the main speed bottleneck.
(b) \textbf{Redundant irrelevant patches}: 
For example, in a randomly selected WSI, numerous instances have extremely low attention scores~\cite{ABMIL}, indicating their minimal contribution, if any, to the bag-level classification.
}
\label{fig:teaser}
\end{figure} 

Recently, multi-instance learning (MIL) has emerged as the leading approach for analyzing pathological whole slide images (WSIs), demonstrating significant success in tasks such as tumor detection, subtyping~\cite{CLAM, DSMIL, DTFD-MIL, TransMIL, S4MIL, WENO}, tissue micro-environment quantification~\cite{schapiro2017histocat, moen2019deep, mahmood2019deep, graham2019hover, saltz2018spatial, javed2020cellular}, and survival prediction~\cite{MambaMIL, BDOCDX, chen2021whole, yao2020whole}. 

To handle gigapixel WSIs, the MIL framework treats each WSI as a bag, cropping it into thousands of patches, each treated as an instance. 
Before being fed into the MIL networks for classification, all patches need to undergo feature extraction.
Considering that each WSI contains thousands of patches, the process of WSI cropping and feature extraction can be very time-consuming.
As shown in~\cref{fig:teaser}, data pre-processing is the primary speed bottleneck, requiring hundreds of times more time than the MIL classifiers. 
Moreover, WSIs often contain redundant patches with minimal contribution to the bag-level classification. 
For example, by adding up the attention scores of only a small fraction (about 10\%) of the patches in the selected WSI, we can obtain 99\% of the total attention scores.
Therefore, the remaining patches can be safely considered as irrelevant and removed without affecting the performance.

Based on the above analysis, a straightforward idea to reduce the inference time is discarding irrelevant instances based on attention scores.
Unfortunately, existing MIL algorithms need to extract the features of all cropped patches before calculating their attention scores, which brings up the ``chicken and egg" problem.
To accelerate WSI classification, Yu et al. proposed SMT~\cite{SMT}. Instead of cropping each WSI into patches, SMT employs cascading vision transformer (ViT) blocks to gradually search for ``suspicious" areas and ultimately uses only a small area of the entire WSI for classification.
As pointed out by Yu et al., the classification performance of SMT heavily relies on accurately identifying potential tumor areas.
However, the pathological information provided by the low-resolution thumbnails, used as the initial input of SMT, is insufficient, which can easily lead to inappropriate regions of interest being focused.
Consequently, the accumulation of errors results in inferior classification performance of SMT when compared to other non-accelerated MIL methods.

In this paper, we propose a hierarchical distillation multi-instance learning (HDMIL) framework aiming to quickly identify irrelevant patches and thus achieve fast and accurate classification.
During training, instance-level features extracted from all cropped patches in the high-resolution WSIs are leveraged to train a dynamic multi-instance network (DMIN) with a self-distillation strategy.
This self-distillation strategy constrains the teacher and student branches in DMIN, which use \textbf{all} and \textbf{partial} instances for classification respectively, to obtain consistent results, thus making the student branch selected instances non-irrelevant.
Afterwards, we can obtain a binary mask for each instance depending on whether the instance is considered relevant to the slide classification.
The masks are then utilized to guide the training of a lightweight instance pre-screening network (LIPN), which learns to identify the binary relevance of each patches in the corresponding low-resolution WSIs.
During testing, after LIPN indicates irrelevant low-resolution patches, we can determine which high-resolution patches can be skipped, thereby saving inference time.
Furthermore, a Chebyshev-polynomials-based Kolmogorov-Arnold (CKA) classifier is designed for more accurate classification, where learnable activation layers have powerful capabilities.

Overall, this paper makes three key contributions:
\begin{itemize}
    \item 
    This paper offers a crucial insight: eliminating irrelevant instances not only speeds up the inference process but also improves the classification performance. 
    This finding challenges the conventional trade-off between speed and performance and provides valuable inspiration for future research in multi-instance classification.
    \item We are the first to propose and apply the Chebyshev-polynomials-based Kolmogorov-Arnold classifier to computational pathology, which can greatly improve the classification performance.    
    \item Extensive experiments on three public datasets demonstrate the effectiveness of our method. For example, on the Camelyon16 dataset, HDMIL achieves an AUC of 90.88\% and an accuracy of 88.61\%, outperforming previous best methods by 3.13\% and 3.18\%, respectively. Moreover, the inference time was reduced by 28.6\%.
\end{itemize}

\section{Related Work}
\label{sec:related}
\noindent\textbf{MIL for WSI Classification.}
MIL for WSI classification can be divided into two categories: instance-based and embedding-based. 
Instance-based methods~\cite{zhu2017deep, pinheiro2015image, feng2017deep, kraus2016classifying, maron1997framework, keeler1990integrated, ramon2000multi} first classify each instance and then aggregate the predictions using Max-Pooling, Mean-Pooling, or other pre-defined pooling operations to generate the final bag-level prediction. 
Embedding-based methods~\cite{ABMIL, CLAM, DSMIL, DTFD-MIL, TransMIL, S4MIL, MambaMIL} use networks to assess the significance of each instance and weight all instances accordingly, producing the bag-level representation for classification.
For the embedding-based methods, it is observed that different instances within each WSI have varying contributions to the bag-level representation. 
Building on this observation, we design the HDMIL framework to achieve fast and accurate classification by selectively removing irrelevant instances.

\noindent\textbf{Dynamic Neural Networks.}
Dynamic neural networks~\cite{han2021dynamic, li2021dynamic, li2020learning, bolukbasi2017adaptive, han2023dynamic, han2022learning, teerapittayanon2016branchynet, yang2020resolution, han2024latency, GumbelSigmoid} can adjust their architecture dynamically according to the input data, thereby controlling the computational redundancy adaptively. 
In the era of Visual Transformers, many studies~\cite{DyT, liang2022not, meng2022adavit, rao2021dynamicvit, song2021dynamic, wang2021not} have attempted to improve inference efficiency by reducing token redundancy. 
In addition to bridging the gap in the field of computational pathology by utilizing dynamic networks to reduce instances and speed up inference, our HDMIL also addresses the aforementioned ``chicken or egg" problem. 
This problem cannot be resolved using existing dynamic networks that solely rely on end-to-end training.

\noindent\textbf{Kolmogorov-Arnold Networks.}
Most previous studies~\cite{sprecher2002space, koppen2002training, lin1993realization, lai2021kolmogorov, leni2013kolmogorov, fakhoury2022exsplinet} before KAN~\cite{KAN} used the original 2-layer structure to explore the possibility of constructing neural networks based on the Kolmogorov-Arnold representation theorem.  
KAN extended this theorem to networks of arbitrary width and depth, exploring its potential as a fundamental model of “AI+Science”.
Subsequent research has primarily focused on improving the integration of KAN into various tasks~\cite{CoxKAN, genet2024temporal, vaca2024kolmogorov, UKAN, TKAN, xu2024effective} or modify its architecture~\cite{xu2024fourierkan, aghaei2024rkan, ConvKAN, yang2024kolmogorov, bozorgasl2024wav, ta2024bsrbf}.
In this paper, we propose to replace the spline function in KAN with first-kind Chebyshev polynomials to develop a more powerful MIL classifier for real-world pathological image classification.

\section{Method}
\label{sec:method}

\begin{figure*}[!t]
    \centering
    \includegraphics[width=1.0\textwidth]{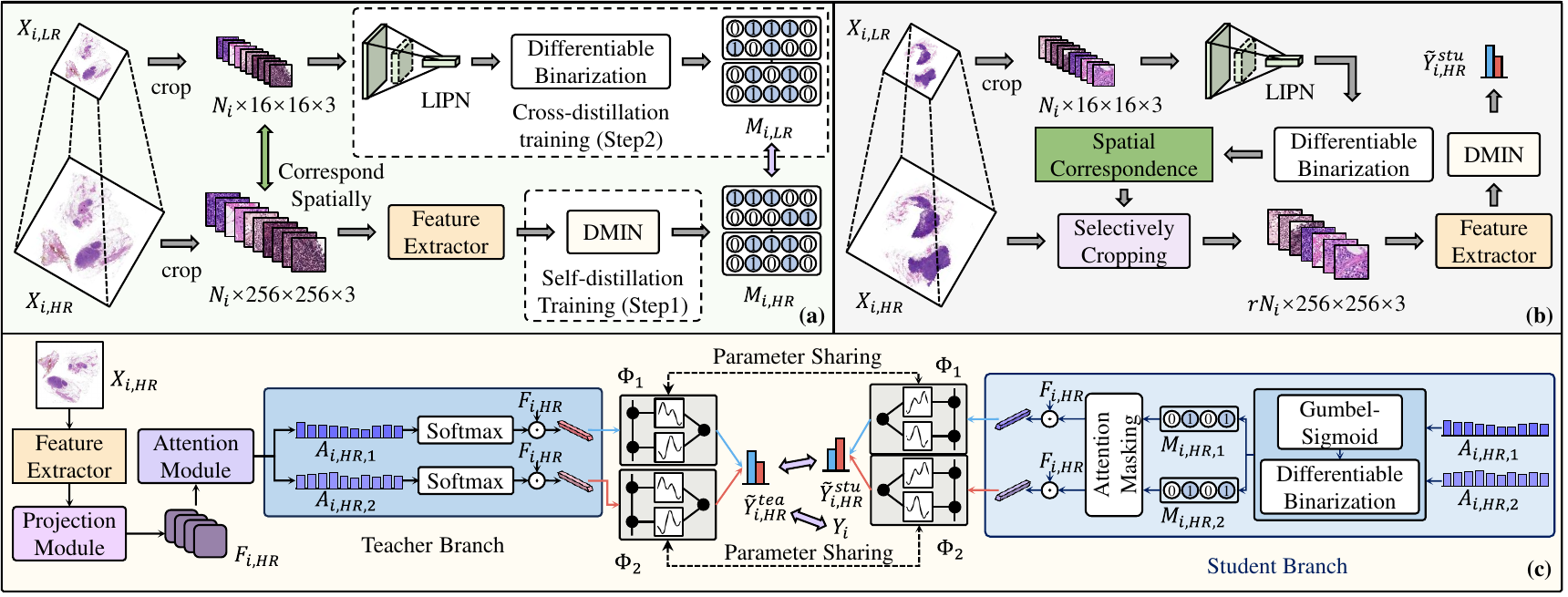}
    \caption{
    Overview of our \textbf{HDMIL} framework. 
    (a) During training, we start by utilize the high-resolution WSI $X_{i,HR}$ for self-distillation of DMIN, enabling it to classify $X_{i,HR}$ and generate per-instance mask $M_{i,HR}$ which indicates the relevance of each region to the bag-level classification. 
    Afterwards we froze DMIN and employ the masks $M_{i,HR}$ to distill LIPN, which learns the contribution of each region using the low-resolution $X_{i,LR}$.
    (b) During inference, the LIPN can identify which patches within $X_{i,HR}$ need to be used for classification by evaluating $X_{i,LR}$.
    (c) The self-distillation training of DMIN on the high-resolution $X_{i,HR}$.
    }
    \label{fig:framework}
\end{figure*}

As illustrated in~\cref{fig:framework}, our proposed HDMIL framework mainly consists of two stages: training and inference.
As shown in~\cref{fig:framework}(a), in the training stage, we first employ a \textbf{self-distillation training} strategy to train the DMIN on high-resolution WSIs for bag-level classification and indicating irrelevant regions.
With the guidance from the trained DMIN, we perform \textbf{cross-distillation training} to get LIPN using low-resolution WSIs, which achieves discrimination of the binary importance (important or not) of each region with extremely low computational cost.
In the inference stage, as shown in~\cref{fig:framework}(b), LIPN relies on low-resolution WSIs to quickly identify regions that are irrelevant to classification and discard the corresponding patches within high-resolution WSIs.
Subsequently, the remaining patches are fed into the feature extractor and DMIN to generate the classification results.

Before training, we first pre-process the input data following the standard procedure for pathological WSIs~\cite{CLAM}.
The dataset $\{X_{i}\}^{S}_{i=1}$ comprises $S$ WSI pyramids with slide labels, where each $X_i$ contains a pair of high-resolution (20$\times$) and low-resolution (1.25$\times$) WSIs, respectively referred to as $X_{i,HR}$ and $X_{i,LR}$.
It should be noted that WSI pyramids typically contain WSIs at various magnification levels ranging from 1.25$\times$ to 40$\times$, but in this paper only the two representative magnifications are utilized.
After removing the background regions, we get $N_i$ pairs of 16$\times$16 patches from $X_{i,LR}$ and 256$\times$256 patches from $X_{i,HR}$.

\subsection{Self-Distillation Training of DMIN}
As shown in~\cref{fig:framework}(c), DMIN is designed to classify high-resolution WSIs and identify instances irrelevant to the bag-level classification.
Specifically, DMIN comprises five modules, namely, the projection module, attention module, teacher branch, student branch, and CKA classifiers.

\noindent\textbf{Projection and Attention Module.}
During training, all patches extracted from the high-resolution WSI $X_{i,HR}$ are fed into a pre-trained feature extractor to generate a set of instance-level features $I_{i,HR}$.
Subsequently, $I_{i,HR}$ is fed into the projection module for dimensionality reduction, producing a new feature set $F_{i,HR} \in \mathbb{R}^{N_i \times Q}$, where $Q$ denotes the dimensionality of the reduced features. 
Then, the dimension-reduced $F_{i,HR}$ is fed into the attention module to compute the un-normalized attention scores:
\begin{equation}
    A_{i,HR} = [\phi({F_{i,HR}}V) \odot \sigma ({F_{i,HR}}U)]W,
\label{eq:attention}
\end{equation}
where $\phi(\cdot)$ and $\sigma(\cdot)$ denote the tanh and sigmoid function.
The weight matrices $U$, $V$, and $W$ are the learnable parameters.
The attention module here uses the same dual branch attention network as CLAM~\cite{CLAM}. 
In the binary classification tasks discussed in this paper, the attention matrices corresponding to the first and second categories are denoted as $A_{i,HR,1}\in\mathbb{R}^{N_{i}\times1}$ and $A_{i,HR,2}\in\mathbb{R}^{N_{i}\times1}$, respectively.

\noindent\textbf{Teacher Branch.}
The dimension-reduced $F_{i,HR}$ is then linearly weighted by the attention matrix for each category to produce the bag-level representation, which are used for final classification:
\begin{equation}
    E_{i,HR,c}^{tea} = \varphi({A_{i,HR,c}})^{\top}\otimes{F_{i,HR,c}}, c\in\{1,2\}.
\label{eq:bag-level representation}
\end{equation}
Here $\varphi(\cdot)$ represents the softmax function and $E_{i,HR,c}^{tea}\in\mathbb{R}^{1\times Q}$ denotes the bag-level representation corresponding to the $c$-th category in the teacher branch.

\noindent\textbf{Student Branch and Self-Distillation.} 
The student branch is designed to compute bag-level representations using only a \textbf{subset} of instances with larger attention scores, and we impose a constraint to ensure that the bag-level representations in the student branch remain as consistent as possible with the representations obtained in the teacher branch using \textbf{all} the instances.
In this way, the attention module is encouraged to focus more on instances that are important for bag-level classification and filters out irrelevant instances.

However, directly using instances with high attention scores is a discrete operation, resulting in a non-differentiable problem during optimization.
To address this issue, we employ the Gumbel trick~\cite{rao2021dynamicvit} to selectively choose instances with higher attention scores for end-to-end training.
First, we incorporate the Gumbel Noise~\cite{GumbelSigmoid} to ``sigmoid" the un-normalized attention matrices:
\begin{equation}
    \hat{A}_{i,HR,c} = \sigma(\frac{{A}_{i,HR,c}+G_{1,c}-G_{2,c}}{\tau}), c\in\{1,2\}.
\label{eq:gumbel-sigmoid}
\end{equation}
Here $\sigma$ represents the sigmoid function, $G_{1,c} \in \mathbb{R}^{N_i \times 1}$ and $G_{2,c} \in \mathbb{R}^{N_i \times 1}$ are two noises matrices randomly sampled from the Gumbel distribution, and $\tau$ is the temperature coefficient.
Next, we binarize the ``sigmoided" attention scores in a differentiable way:
\begin{equation}
    {M}_{i,HR,c}^{j} = B(\hat{A}_{i,HR,c}^{j},\gamma)-D(\hat{A}_{i,HR,c}^{j})+\hat{A}_{i,HR,c}^{j},
\label{eq:discrete-mask}
\end{equation}
where ${M}_{i,HR,c}^{j}\in\{0,1\}$ represents the mask value of the $j$-th instance and $\gamma$ denotes the threshold as a hyper-parameter.
$B(a,b)$ here represents the discrete binarization function, which equals $1$ when $a$ is greater than $b$, and $0$ otherwise.
$D(\cdot)$ represents the gradient truncation operation.

Furthermore, we propose an \textbf{attention masking} mechanism to eliminate the impact of instances with zero mask values on the bag-level representations:
\begin{equation}
    E_{i,HR,c}^{stu} = \sum_{j=1}^{N_i}\frac{exp({A}_{i,HR,c}^{j}){M}_{i,HR,c}^{j}}{\sum_{s=1}^{N_i}exp({A}_{i,HR,c}^{s}){M}_{i,HR,c}^{s}}F_{i,HR,c}^{j},
\label{eq:attention_mask}
\end{equation}
where $E_{i,HR,c}^{stu}\in \mathbb{R}^{1\times Q}$ represents the bag-level representation of the $c$-th class in the student branch.

\noindent\textbf{CKA Classifier.} 
In order to enhance the capacity of the MIL classifier, we propose to use the Kolmogorov-Arnold network to learn nonlinear activation functions instead of using fixed activation functions in the classifier. 
Specifically, we employ the iterative form of $K$-order Chebyshev polynomials to represents the basis functions $T_{K}(x)$:
\begin{equation}
    T_{K}(x) = 2xT_{K-1}(x) - T_{K-2}(x), K\ge{2}.
\label{eq:chebyshev-polynomial}
\end{equation}
Here $x\in\mathbb{R}^{1\times Q}$ represents a bag-level representation, where the baseline condition are $T_{0}(x)=\vec{\mathbf{1}}$ and $T_{1}(x)= x$. By multiplying the basis functions $T(x)$ by the learnable coefficients $\Omega\in\mathbb{R}^{Q\times O\times (K+1)}$, we can get the prediction of the classifier $\Phi(x)$:
\begin{equation}
    \Phi(x)[o] = \sum_{k=0}^{K}\sum_{q=1}^{Q}T_{k}(x)[q] * \Omega[q, o, k],
\label{eq:CKA}
\end{equation}
where $O$ represents the dimension of the prediction result.
Since we use a dual-branch attention module, it's natural to calculate the classification results for the two branches individually, so $O$ is equal to 1, and the predictions of the teacher branch and the student branch are:
\begin{equation}
\begin{cases}
    \widetilde{Y}_{i,HR}^{tea} = [\Phi_{1}(\phi(E_{i,HR,1}^{tea}))\oplus\Phi_{2}(\phi(E_{i,HR,2}^{tea}))]. \\
    \widetilde{Y}_{i,HR}^{stu} = [\Phi_{1}(\phi(E_{i,HR,1}^{stu}))\oplus\Phi_{2}(\phi(E_{i,HR,2}^{stu}))].
\label{eq:prediction}
\end{cases}
\end{equation}
Here $\oplus$ represents the concatenation operation and the tanh function $\phi$ maps the input values of the CKA classifiers to $\left[-1,1\right]$, ensuring that the inputs meet the requirements of the Chebyshev polynomial.

\noindent\textbf{Hybrid Loss Function.}
The training objectives of DMIN are threefolds: 
1) The teacher branch can correctly classify $X_{i,HR}$; 
2) The classification results of the student branch (using partial instances) and teacher branch (using all instances) should be consistent; 
3) The proportion of instances selected should be controllable. 
Specifically, we first use the cross-entropy loss $L_{cls}^{tea}$ to ensure that the teacher branch performs accurate classification:
\begin{equation}
    L_{cls}^{tea} = CE(\widetilde{Y}_{i,HR}^{tea}, Y_i),
\end{equation}
where $CE(\cdot)$ represents the cross entropy loss function and $Y_i$ is the slide-level label of $X_i$.
Next, we constrain the bag-level representation $E_{i,HR}^{stu}$ and classification logit $\widetilde{Y}_{i,HR}^{stu}$ in the student branch by knowledge distillation:
\begin{equation}
\begin{cases}
    L_{dis,1}^{stu} = L_{2}(E_{i,HR}^{stu}, E_{i,HR}^{tea}), \\
    L_{dis,2}^{stu} = L_{KL}(\widetilde{Y}_{i,HR}^{stu}, \widetilde{Y}_{i,HR}^{tea}).
\end{cases}
\end{equation}
Here, $L_{2}(\cdot)$ and $L_{KL}(\cdot)$ denote the 2-norm and KL divergence loss function, respectively. 
Finally, we constrain the proportion of learned relevant instances $\widetilde{r}_{i, HR}$ to be close to a preset retention ratio $r$:
\begin{equation}
    L_{rate}^{stu} = L_{2}(\widetilde{r}_{i, HR}, r).
\end{equation}
Here, the $j$-th instance is considered relevant if either ${M}_{i,HR,1}^{j}$ or ${M}_{i,HR,2}^{j}$ are not zero.
Additionally, we utilize the clustering loss $L_{clu}^{tea}$ proposed in CLAM~\cite{CLAM} to optimize the feature space of DMIN.
In conclusion, the hybrid loss function of DMIN is:
\begin{equation}
    L_{DMIN} = \alpha_{1}L_{cls}^{tea} + \alpha_{2}L_{clu}^{tea} + \alpha_{3}L_{dis,1}^{stu}+ \alpha_{4}L_{dis,1}^{stu} + \alpha_{5}L_{rate}^{stu}.
\label{eq:DMIN-loss}
\end{equation}
For the coefficients of different loss terms, we did not perform hyper-parameter search, but empirically set $\alpha_1$ and $\alpha_2$ to 0.7 and 0.3 according to CLAM~\cite{CLAM}, and set $\alpha_3$, $\alpha_4$, and $\alpha_5$ to 0.5, 0.5, and 2.0 according to DynamicViT~\cite{rao2021dynamicvit}.

\subsection{Cross-Distillation Training of LIPN}
Although DMIN can successfully identify irrelevant regions within WSIs, it does not improve the inference speed. 
This is because DMIN needs to use all patches' features generated by the feature extractor to determine which instances should be discarded. 
However, this patch-wise feature extraction is actually the bottleneck for WSI inference speed. 
To solve this problem, we propose using DMIN to distill LIPN, a lightweight instance pre-screening network specifically tailored for low-resolution WSIs, as shown in~\cref{fig:framework}(a). 
After training, LIPN can quickly identify the irrelevant regions within low-resolution WSIs, thereby indirectly indicating the irrelevant patches within high-resolution WSIs.

Specifically, the $N_i$ 16$\times$16 patches obtained from $X_{i,LR}$ are directly fed into LIPN, generating dual-branch prediction matrices $P_{i,LR,c}, c\in\{1,2\}$ for the two categories.
Since these low-resolution patches contain relatively little information, we do not require LIPN to learn the specific contribution score of each patch to the bag-level classification like DMIN does.
On the contrary, it is easier for LIPN to learn whether each patch contributes to the bag-level classification or not.
Therefore, $P_{i,LR,c}$ is first binarized:
\begin{equation}
    {M}_{i,LR,c}^{j} = B(P_{i,LR,c}^{j},\gamma)-D(P_{i,LR,c}^{j})+P_{i,LR,c}^{j}.
\end{equation}
Next, $M_{i,LR,c}$ is forced to be consistent with $M_{i,HR,c}$, so that $M_{i,LR,c}$ can also indicate whether an patch is relevant.
What's more, the ratio of learned relevant patches $\widetilde{r}_{i, LR}$ is also constrained to be close to $r$.
Overall, the hybrid loss function of LIPN is:
\begin{equation}
    L_{LIPN} = \beta_{1}\sum_{c=1}^{2}\frac{L_{1}(M_{i, LR,, c}, M_{i, HR, c})}{2} + \beta_{2}L_{2}(\widetilde{r}_{i, LR}, r).
\end{equation}
Here, $L1(\cdot)$ denotes the 1-norm loss function.
In our implementation, we employed the widely-used ResNet-50 pre-trained on ImageNet as the feature extractor, and used a lightweight variant of MobileNetV4~\cite{MobileNetv4} for the pre-screening network LIPN. The detailed architecture of LIPN is illustrated in the supplementary material.

\subsection{Efficient Inference}
As shown in~\cref{fig:framework}(b), our proposed efficient inference process consists of three steps: 
1) Cropping all patches from $X_{i,LR}$, with the total number of patches being $N_i$.
2) Feeding these patches into LIPN to identify regions relevant to classification, generating $M_{i,LR}$;
3) Selectively cropping relevant $\widetilde{r}_{i, LR}N_i$ patches from $X_{i,HR}$ based on $M_{i,LR}$, and then feeding them into the feature extractor and DMIN.
Afterwards, we calculate the bag-level representations and the final classification results using the student branch across categories separately.

\section{Experimental Results}
\label{sec:experiments}

\begin{table*}[!t]
\setlength{\tabcolsep}{3pt}
\renewcommand{\arraystretch}{1.0}
\centering
\begin{tabular}{l|ccc|ccc|ccc}
\toprule
\multirow{2}{*}{\makecell{Comparative\\Methods}} & \multicolumn{3}{c|}{Camelyon16} & \multicolumn{3}{c|}{TCGA-NSCLC} & \multicolumn{3}{c}{TCGA-BRCA} \\
\cline{2-10}
~ & AUC$\uparrow$ & ACC$\uparrow$ & Time(s)$\downarrow$ & AUC$\uparrow$ & ACC$\uparrow$ & Time(s)$\downarrow$ & AUC$\uparrow$ & ACC$\uparrow$ &Time(s)$\downarrow$ \\
\midrule
Max-Pooling & ${83.26}_{1.54}$ & ${82.41}_{0.73}$ & ${23.46}$  & ${94.66}_{2.33}$ & ${86.40}_{3.73}$ & ${57.16}$ & ${88.03}_{7.76}$ & ${86.05}_{3.88}$ & ${36.49}$  \\
Mean-Pooling  & ${61.80}_{2.15}$ & ${70.54}_{1.41}$ & ${23.46}$  & ${92.82}_{3.54}$ & ${84.93}_{4.78}$ & ${57.16}$  & ${88.23}_{5.67}$ & ${86.74}_{2.44}$ & ${36.49}$  \\
ABMIL~\cite{ABMIL}   & ${84.88}_{3.38}$ & ${82.79}_{2.68}$ & ${23.46}$  & ${94.92}_{2.29}$ & ${88.03}_{3.65}$ & ${57.16}$ & ${87.70}_{6.15}$ & ${87.68}_{3.51}$  & ${36.49}$ \\
CLAMSB~\cite{CLAM}   & ${83.49}_{4.46}$ & ${79.61}_{4.40}$ & ${23.46}$ & ${95.05}_{2.72}$ & ${88.74}_{3.39}$ & ${57.16}$ & ${88.25}_{6.12}$ & ${87.58}_{4.92}$  & ${36.49}$ \\
CLAMMB~\cite{CLAM}   & ${87.51}_{3.23}$ & ${82.56}_{3.11}$ & ${23.46}$ & ${95.59}_{2.16}$ & ${88.01}_{3.38}$ & ${57.16}$ & ${90.22}_{5.18}$ & ${88.27}_{3.52}$  & ${36.49}$ \\
DSMIL~\cite{DSMIL}   & ${75.94}_{10.81}$ & ${75.35}_{6.12}$ & ${23.46}$  & ${92.11}_{2.97}$ & ${83.67}_{3.80}$ & ${57.16}$ & ${83.33}_{7.48}$ & ${82.59}_{3.66}$  & ${36.49}$ \\
TransMIL~\cite{TransMIL}   & ${82.26}_{5.67}$ & ${81.01}_{6.85}$ & ${23.47}$  & ${94.57}_{2.03}$ & ${88.21}_{3.04}$ & ${57.17}$ & ${88.33}_{5.73}$ & ${87.55}_{3.78}$ & ${36.49}$ \\
DTFDAFS~\cite{DTFD-MIL}  & ${87.40}_{3.17}$ & ${85.12}_{2.42}$ & ${23.46}$  & ${95.59}_{2.08}$ & ${88.76}_{3.89}$ & ${57.16}$ & ${87.24}_{7.38}$ & ${86.83}_{3.98}$  & ${36.49}$  \\
DTFDMAS~\cite{DTFD-MIL}  & ${87.75}_{2.07}$ & ${85.43}_{2.03}$ & ${23.46}$  & ${95.02}_{2.32}$ & ${89.02}_{3.78}$ & ${57.17}$ & ${87.80}_{9.65}$ & ${87.48}_{4.13}$ & ${36.49}$  \\
S4MIL~\cite{S4MIL}  & ${86.40}_{1.99}$ & ${80.39}_{2.79}$ & ${23.47}$ & ${96.19}_{1.89}$ & ${89.69}_{2.86}$ & ${57.17}$ & ${90.40}_{5.73}$ & ${88.17}_{3.88}$  & ${36.49}$ \\
MambaMIL~\cite{MambaMIL}   & ${87.06}_{6.19}$ & ${83.26}_{2.93}$ & ${23.47}$  & ${95.37}_{1.70}$ & ${89.62}_{3.13}$ & ${57.16}$ & ${89.69}_{5.91}$ & ${87.78}_{4.27}$  & ${36.49}$ \\
\hline
HDMIL$\dag$  & $\textcolor{red}{\textbf{93.17}}_{1.83}$ & $\textcolor{red}{\textbf{88.92}}_{2.51}$ & $\textcolor{blue}{\textbf{23.46}}$  & $\textcolor{red}{\textbf{96.47}}_{2.20}$ & $\textcolor{blue}{\textbf{89.75}}_{2.86}$ & $\textcolor{blue}{\textbf{57.16}}$ & $\textcolor{blue}{\textbf{90.43}}_{4.86}$ & $\textcolor{red}{\textbf{88.68}}_{3.17}$  & $\textcolor{blue}{\textbf{36.49}}$  \\
HDMIL &  $\textcolor{blue}{\textbf{90.88}}_{2.75}$ & $\textcolor{blue}{\textbf{88.61}}_{2.04}$ & $\textcolor{red}{\textbf{16.75}}$  & $\textcolor{blue}{\textbf{96.35}}_{2.26}$ & $\textcolor{red}{\textbf{89.78}}_{3.11}$  & $\textcolor{red}{\textbf{44.71}}$ & $\textcolor{red}{\textbf{90.45}}_{4.42}$ & $\textcolor{blue}{\textbf{88.27}}_{2.47}$  & $\textcolor{red}{\textbf{33.86}}$  \\
\bottomrule
\end{tabular}
\caption{
Comparison of HDMIL with the state-of-the-art MIL methods on Camelyon16, TCGA-NSCLC, and TCGA-BRCA. 
The 10-fold \textbf{test} AUC and accuracy (ACC) scores are reported in the form of $\mathrm{{mean}_{std}}$.
The best and second best results are indicated in red and blue, respectively.
The average processing time per WSI on each test sets are also shown.
HDMIL$\dag$ means using only DMIN for inference.
}
\label{tab:comparative_results}
\end{table*}

\begin{table}[!t]
\setlength{\tabcolsep}{0.5pt}
\renewcommand{\arraystretch}{1.0}
\centering
\begin{tabular}{l|c|cccc|c}
\toprule
Methods & Dataset & LIPN & Crop & Fea & DMIN & Total \\
\midrule
\multirow{3}{*}{Came16} & HDMIL$\dag$ & - &  $13.45$ & $10.00$ & $0.02$  & $23.46$ \\
~ & HDMIL & $0.01$ & $10.88$ & $5.84$ & $0.02$  & $16.75$ \\
~ & $\Delta$ & - & $-{\textbf{19.1}}\%$ & $-{\textbf{41.6}}\%$ & -  & $-{\textbf{28.6}}\%$ \\
\hline
\multirow{3}{*}{NSCLC} & HDMIL$\dag$ & - &  $47.02$ & $10.12$ & $0.02$  & $57.16$  \\
~ & HDMIL &$0.01$  & $37.21$ & $7.48$ & $0.02$ & $44.71$ \\
~ & $\Delta$ & - & $-{\textbf{20.9}}\%$ & $-{\textbf{26.1}}\%$  & - & $-{\textbf{21.8}}\%$ \\
\hline
\multirow{3}{*}{BRCA} & HDMIL$\dag$ & - &  $27.17$ & $9.30$ & $0.02$  & $36.49$ \\
~ & HDMIL  &$0.01$  & $25.84$ & $8.00$ & $0.02$ & $33.86$ \\
~ & $\Delta$ & - & $-{\textbf{4.90}}\%$ & $-{\textbf{14.0}}\%$  & - & $-{\textbf{7.2}}\%$ \\
\bottomrule
\end{tabular}
\caption{
Comparison of HDMIL and HDMIL$\dag$ when splitting the inference time (seconds) into four stages: instance pre-screening (LIPN), WSI cropping (``Crop"), feature extraction (``Fea"), and bag classification (DMIN).
}
\label{tab:process_time}
\end{table}

\subsection{Settings}
We evaluated our proposed algorithm on three public datasets:
1) for breast cancer lymph node metastasis detection using the Camelyon16~\cite{Camelyon16} dataset; 
2) for lung cancer subtyping using the TCGA-NSCLC dataset; 
and 3) for breast cancer subtyping using the TCGA-BRCA dataset. 
All WSIs were pre-processed using tools developed by CLAM~\cite{CLAM}. 
All experiments adhered to the principle of \textbf{10-fold Monte Carlo cross-validation}. 
For Camelyon16, the official training set was divided into training and validation sets at a 9:1 ratio based on the number of cases in each fold, while the official test set was used for testing across all folds. 
The TCGA-NSCLC and TCGA-BRCA datasets were split into the training, validation, and test sets in an 8:1:1 ratio, again based on the number of cases in each fold.
The implementation details are in the supplementary material.

\subsection{Comparative Results on Test Sets}
\noindent\textbf{Classification Performance.}
\Cref{tab:comparative_results} compares the classification performance of our proposed HDMIL against existing MIL methods on the Camelyon16~\cite{Camelyon16}, TCGA-NSCLC, and TCGA-BRCA test sets.
HDMIL$\dag$ means using only DMIN for inference without pre-screening instances through LIPN.
From the table we can find: 1) Both HDMIL$\dag$ and HDMIL consistently outperform existing methods across these datasets.
2) When the dataset is large enough, the speedup brought by HDMIL does not means a decrease in classification performance. For example, the test performance gap between HDMIL$\dag$ and HDMIL is small on TCGA-NSCLC and TCGA-BRCA, both of which contain about 1000 WSIs.
Meanwhile, the AUC score of HDMIL decreases slightly on Camelyon16, but is still much better than existing MIL methods.
We believe that the performance degradation of HDMIL compared to HDMIL$\dag$ on the Camelyon16 dataset can be primarily attributed to the small dataset size (less than 400 WSIs), rather than inherent shortcomings of HDMIL itself. 
A further analysis is presented in~\cref{subsec:further analysis}.

\noindent\textbf{Inference Time.}
From~\cref{tab:comparative_results}, it is evident that the processing time of HDMIL$\dag$ is nearly identical to that of existing methods since they need to \textbf{process the same number of high-resolution patches}. However, HDMIL outperforms all other methods, significantly reducing the processing time.
Compared to HDMIL$\dag$, HDMIL achieves an total speed improvement of 28.6\%, 21.8\%, and 7.2\% on the three datasets, respectively.
To analyze how HDMIL achieves this time-saving effect, we divide the WSI inference process into four stages: instance pre-screening, WSI cropping, feature extraction, and MIL classification, as presented in~\cref{tab:process_time}.
Although LIPN causes a slight increase in inference time (approximately 0.01 seconds), it reduces the number of instances that require cropping and feature extraction, thereby significantly reducing total inference time.

\begin{figure}[!h]
  \centering
  \includegraphics[width=0.48\textwidth]{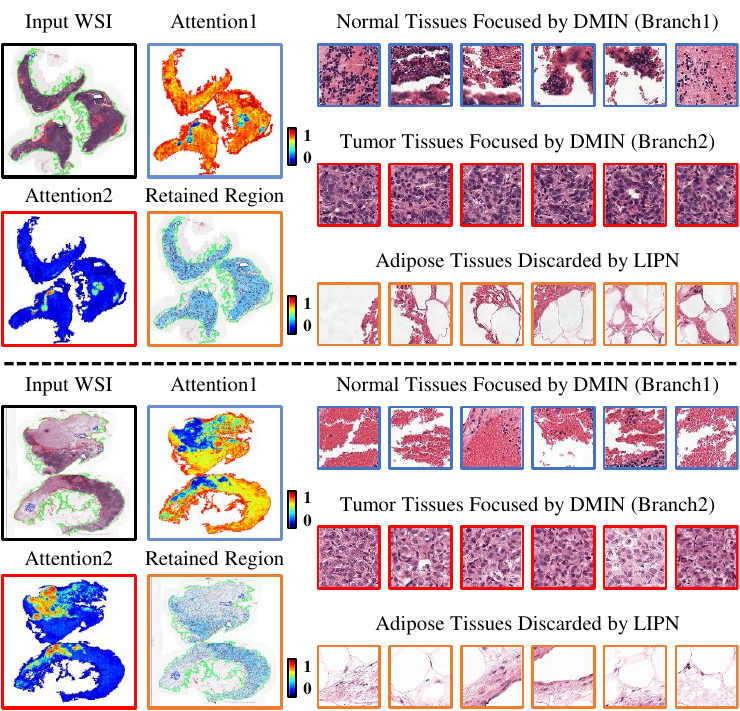}
  \caption{Visualization analysis of two randomly selected WSIs. The pathologists marked the tumor areas in the input WSIs with red lines.
  The dual-branch attention maps in DMIN (``Attention1" and ``Attention2") are shown, and the instances selected by LIPN are marked with  blue masks (``Retained Region")
  }
  \label{fig:visualization}
\end{figure}

\begin{table*}[!t]
\setlength{\tabcolsep}{5.5pt}
\renewcommand{\arraystretch}{1.0}
\centering
\begin{tabular}{c|c|c|cc|cc|cc|cc}
\toprule
\multicolumn{2}{c|}{DMIN} & \multirow{2}{*}{LIPN} & \multicolumn{2}{c|}{Camelyon16} & \multicolumn{2}{c|}{TCGA-NSCLC} & \multicolumn{2}{c|}{TCGA-BRCA} & \multicolumn{2}{c}{Average} \\
\cline{1-2}
\cline{4-11}
CKA & SelfDist & ~ & AUC & ACC & AUC & ACC & AUC & ACC & AUC & ACC \\
\midrule
\ding{55} & \ding{55} & \ding{55} & ${94.67}_{4.51}$ & ${91.54}_{5.38}$ & ${95.36}_{3.51}$ & ${89.44}_{4.51}$ & ${88.82}_{6.41}$ & ${86.47}_{4.21}$ & ${92.95}$ & ${89.15}$ \\
\ding{51} & \ding{55} & \ding{55} & ${97.15}_{3.27}$ & ${93.85}_{4.86}$ & ${95.19}_{3.01}$ & ${89.67}_{3.57}$ & ${91.22}_{5.40}$ & $\textcolor{blue}{\textbf{89.00}}_{3.62}$ & ${94.52}$ & ${90.84}$ \\
\ding{51} & \ding{51} & \ding{55} & $\textcolor{red}{\textbf{97.70}}_{2.54}$ & $\textcolor{blue}{\textbf{95.00}}_{4.81}$ & $\textcolor{blue}{\textbf{95.58}}_{3.27}$ & $\textcolor{blue}{\textbf{90.29}}_{3.90}$ & $\textcolor{red}{\textbf{93.33}}_{4.58}$ & $\textcolor{red}{\textbf{89.83}}_{2.71}$ & $\textcolor{blue}{\textbf{95.54}}$ & $\textcolor{red}{\textbf{91.71}}$ \\
\ding{51} & \ding{51} & \ding{51} & $\textcolor{blue}{\textbf{97.64}}_{2.93}$ & $\textcolor{red}{\textbf{95.38}}_{3.97}$ & $\textcolor{red}{\textbf{95.88}}_{3.02}$ & $\textcolor{red}{\textbf{90.50}}_{3.44}$ & $\textcolor{blue}{\textbf{93.27}}_{4.87}$ & ${88.70}_{3.92}$ & $\textcolor{red}{\textbf{95.60}}$ & $\textcolor{blue}{\textbf{91.53}}$ \\
\bottomrule
\end{tabular}
\caption{
The effect of each component in HDMIL on classification performance.
The 10-fold \textbf{validation} AUC and ACC scores are reported in the form of $\mathrm{{mean}_{std}}$.
``SelfDist" is the abbreviation for self-distillation.
}
\label{tab:each_component}
\end{table*}

\renewcommand\thesubtable{\arabic{subtable}}
\begin{table}[!t]
\setlength{\tabcolsep}{8pt}
\renewcommand{\arraystretch}{1.0}

\begin{subtable}[b]{1.0\linewidth}
\centering
\setlength{\tabcolsep}{11pt}
\begin{tabular}{c|ccc}
\toprule
& Projection & Attention  & Classifier \\
\midrule
Params & 7.082M & 3.942M & 0.828M \\
\hline
AUC & $\textcolor{blue}{\textbf{92.79}}_{6.92}$ & ${85.03}_{8.36}$ & $\textcolor{red}{\textbf{97.15}}_{3.27}$  \\
ACC & $\textcolor{blue}{\textbf{86.54}}_{9.11}$ & ${77.31}_{6.40}$ & $\textcolor{red}{\textbf{93.85}}_{4.86}$ \\
\bottomrule
\end{tabular}
\captionsetup{size=normal}
\caption{Comparison of the position of the CKA layer.}
\label{tab:speed_compare1}
\end{subtable}

\begin{subtable}[b]{1.0\linewidth}
\centering
\setlength{\tabcolsep}{5.3pt}
\begin{tabular}{c|cccc}
\toprule
& FC & MLP  & KA~\cite{KAN} & CKA \\
\midrule
Params & 0.791M & 1.842M & 0.828M & 0.828M\\
\hline
AUC & ${94.67}_{4.51}$ & ${94.97}_{5.05}$ & $\textcolor{blue}{\textbf{96.42}}_{2.88}$ & $\textcolor{red}{\textbf{97.15}}_{3.27}$  \\
ACC & ${91.54}_{5.38}$ & $\textcolor{blue}{\textbf{92.69}}_{6.13}$ & ${91.16}_{6.03}$ & $\textcolor{red}{\textbf{93.85}}_{4.86}$ \\
\bottomrule
\end{tabular}
\captionsetup{size=normal}
\caption{Comparison of FC, MLP, KA, and CKA as classifiers.}
\label{tab:speed_compare1}
\end{subtable}

\begin{subtable}[b]{1.0\linewidth}
\centering
\setlength{\tabcolsep}{5.2pt}
\begin{tabular}{c|cccc}
\toprule
& K=4 & K=8 & K=12  & K=16 \\
\midrule
Params & 0.803M & 0.816M & 0.828M & 0.840M \\
\hline
AUC & ${94.67}_{4.29}$ & ${94.61}_{4.48}$ & $\textcolor{red}{\textbf{97.15}}_{3.27}$  & $\textcolor{blue}{\textbf{96.18}}_{2.59}$ \\
ACC & ${89.62}_{7.91}$ & ${90.38}_{7.08}$ & $\textcolor{red}{\textbf{93.85}}_{4.86}$  & $\textcolor{blue}{\textbf{90.39}}_{6.60}$ \\
\bottomrule
\end{tabular}
\captionsetup{size=normal}
\caption{Comparison of using different orders in CKA classifier.}
\label{tab:speed_compare1}
\end{subtable}
\caption{
Analysis of the CKA classifier. 
The 10-fold \textbf{validation} performance on the Camelyon16 dataset are reported.
}
\label{tab:CKA_comaprsion}
\end{table}

\subsection{Focusing and Discarding Visualization}

\Cref{fig:visualization} shows two tumor WSIs with patch-level annotations, attention maps generated by DMIN, the instance retention after LIPN pre-screening, DMIN-focused patches, and LIPN-discarded patches.
As expected, the first branch of DMIN focuses on normal tissue regions, while the second branch emphasizes tumor regions, demonstrating DMIN’s ability to identify  regions related to bag-level classification.
Moreover, the regions to which DMIN assigns greater importance are retained by LIPN, while instances derived from adipose tissues, which contribute minimally to classification, are effectively discarded by LIPN.

\subsection{Ablation Study on Validation Sets}
\noindent\textbf{Effect of Each Components.}
\Cref{tab:each_component} presents the impact  of each module in HDMIL on the classification results. 
Notably, replacing the conventional linear layer-based classifier with the proposed CKA classifiers and incorporating self-distillation into the DMIN training, both significantly improve the classification performance. 
In addition, using LIPN for instance pre-screening does not result in a obvious decrease in the classification performance on the validation set, slightly differing from the situation on the test set in~\cref{tab:comparative_results}.
This will also be discussed in~\cref{subsec:further analysis}.
In the following subsections, we analyze the reasons why each component works by 10-fold cross-validation experiments.

\noindent\textbf{CKA Classifier in DMIN.}
As shown in~\cref{tab:CKA_comaprsion}, we analyze the proposed CKA classifiers from three perspectives: 
1) the impact of employing the CKA layer at different positions; 
2) comparison with other classification layers; 
and 3) the impact of different Chebyshev polynomial orders $K$. 
To eliminate the impact of other factors, all experiments here only utilize the teacher branch trained on all instances.
\begin{itemize}
    \item When using CKA layers as the projection or attention module, the number of trainable parameters increases dramatically, accompanied by a significant drop in performance. 
    It seems that our CKA layer is also better at solving modeling problems in lower dimensional spaces, similar to the vanilla KAN~\cite{KAN}.
    \item When compared with other classifiers such as the FC layer, two-layer MLP, and KA~\cite{KAN} layer, CKA demonstrates superior classification performance.
    Although FC, MLP, and KA can sometimes achieve similar performance to CKA in certain folds, there tends to be a larger performance gap in other folds.
    Thus, CKA is a more powerful and robust classifier.
    \item When the Chebyshev polynomial order changes from 4 to 16, the number of parameters of the entire DMIN does not change much. Nevertheless, there is a noticeable disparity in classification performance, with the best outcome achieved at an order of 12. Further increasing the order does not lead to better improvements in classification performance, probably due to the limited training data.
\end{itemize}

\noindent\textbf{Self-Distillation of DMIN.}
We believe that self-distillation enhances the classification performance by enforcing the attention module to focus on crucial instances, thereby reducing the impact of irrelevant regions. This can be seen as a form of ``denoising".
To verify this viewpoint, we evaluated the quality of the bags after the “denoising” effect of self-distillation by considering three types of instances to represent each bag: all instances within each WSI, instances selected by the trained DMIN, and randomly sampled instances.   
Newly trained Max-Pooling models are used to evaluate the quality of these three types of bags like linear probing~\cite{SimCLR}.
As shown in~\cref{tab:line_prob}, MIL models trained with instances selected by DMIN outperform models trained with randomly sampled instances and even outperform models trained with all instances. 
This suggests that self-distillation \textbf{improves the quality of bags} for classification by instance selection.

\begin{table}[!h]
\setlength{\tabcolsep}{4pt}
\renewcommand{\arraystretch}{1.0}
\centering
\begin{tabular}{c|c|c|c}
\toprule
Metrics & All & Random & DMIN \\
\midrule
AUC/ACC & $\textcolor{blue}{\textbf{92.06}}/\textcolor{blue}{\textbf{84.62}}$ & ${89.40}/{83.46}$ & $\textcolor{red}{\textbf{92.73}}/\textcolor{red}{\textbf{85.78}}$ \\
\bottomrule
\end{tabular}
\caption{
Average performance of Max-Pooling trained with different kinds of bags on the Camelyon16 \textbf{validation} set. The number of instances in each ``random" bag were kept consistent with the number of instances in each ``DMIN" bag.
}
\label{tab:line_prob}
\end{table}

\noindent\textbf{Distillation Methods in LIPN.}
\Cref{tab:distillation_comparison} explores the effects of different distillation methods when using DMIN to distill LIPN.
The symbol ${A}_{H}\to{P}_{L}$ represents the distillation from attention $A_{i,HR}$ to instance-wise predictions $P_{i,LR}$, while ${M}_{H}\to{M}_{L}$ denotes the distillation between $M_{i,HR}$ and $M_{i,LR}$.
It can be seen that distilling among discrete masks yields significantly better results, especially on the Camelyon16 dataset. 
This is because low-resolution patches lose too much information, in which case learning to predict the specific contribution score of each instance becomes challenging for LIPN, compared to learning the binary decision of the instance (keep or discard).

\begin{table}[!h]
\setlength{\tabcolsep}{1.6pt}
\renewcommand{\arraystretch}{1.0}
\centering
\begin{tabular}{c|cc|cc|cc}
\toprule
\multirow{2}{*}{\makecell{Distillation\\Manner}} & \multicolumn{2}{c|}{Camelyon16} & \multicolumn{2}{c|}{TCGA-NSCLC} & \multicolumn{2}{c}{TCGA-BRCA} \\
\cline{2-7}
~ & AUC & ACC & AUC & ACC & AUC & ACC \\
\midrule
${A}_{H}\to{P}_{L}$ & ${81.70}$  & ${83.85}$ & ${95.01}$  & ${89.29}$ & ${90.49}$  & ${87.10}$ \\
${M}_{H}\to{M}_{L}$ & $\textbf{97.64}$  & $\textbf{95.38}$ & $\textbf{95.88}$  & $\textbf{90.50}$ & $\textbf{93.27}$  & $\textbf{88.70}$ \\
\bottomrule
\end{tabular}
\caption{
Performance of HDMIL on the \textbf{validation} set when using different distillation methods.
${A}_{H}$, ${P}_{L}$, ${M}_{H}$, and ${M}_{L}$ are the abbreviations for ${A}_{i, HR}$, ${P}_{i, LR}$, ${M}_{i, HR}$, and ${M}_{i, LR}$ respectively.
}
\label{tab:distillation_comparison}
\end{table}

\begin{figure}[!h]
  \centering
  \includegraphics[width=0.48\textwidth]{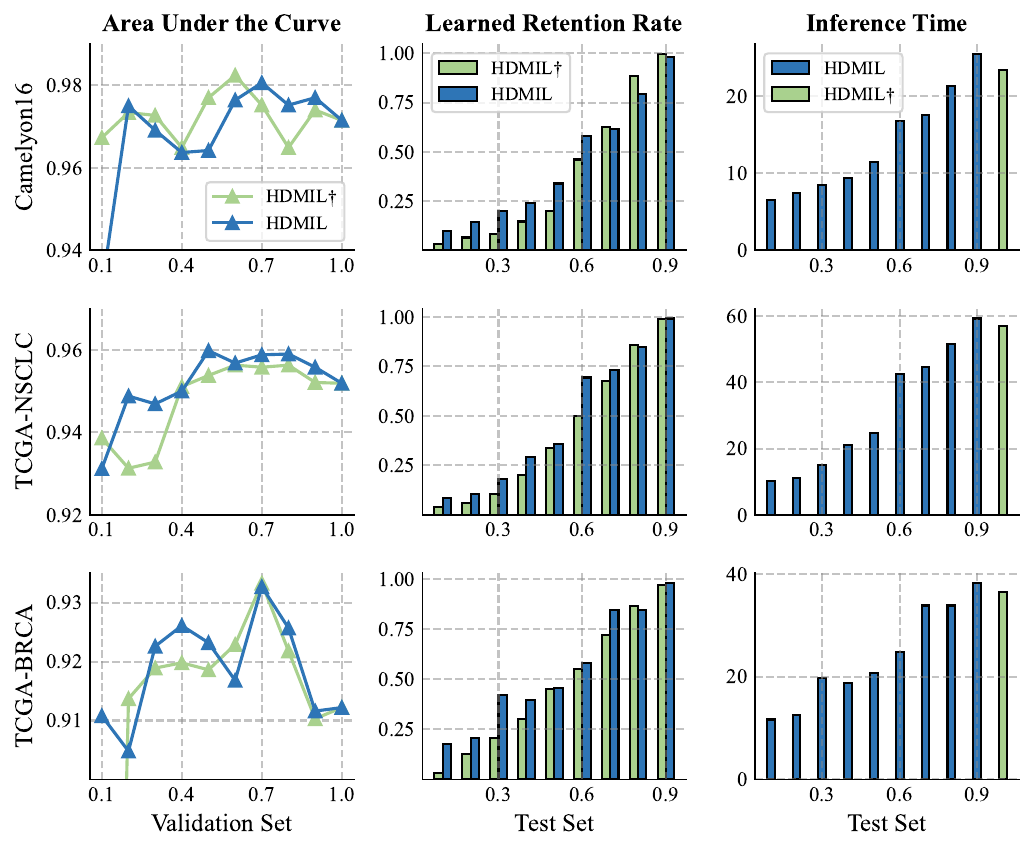}
  \caption{
  The impact of the preset instance retention rate $\mathbf{r}$ (hyper-parameter) on classification performance, actual learned instance retention ratio, and inference time (seconds).
  }
  \label{fig:mask_ratio}
\end{figure}

\noindent\textbf{Impact of the Preset Instance Retention Ratio.}
We examine the impact of the preset instance ratio $r$ on the classification performance, actual learned instance retention ratios, and inference time, as depicted in~\cref{fig:mask_ratio}: 1) Generally, the performance of HDMIL and HDMIL$\dag$ exhibits a pattern of initial improvement followed by a decline as $r$ increases.
This can be attributed to the fact that when $r$ is small, the scarcity of patches leads to the loss of classification-related information. Conversely, when $r$ becomes excessively large, the self-distillation training fails to eliminate the interference caused by irrelevant instances.
2) In addition, it can be seen that the instance retention rate actually learned by HDMIL$\dag$ and HDMIL is roughly equivalent to the preset $r$.
3) The inference time gradually increases as $r$ increases. When $r$ reaches 0.9, the total inference time of HDMIL surpasses that of HDMIL$\dag$ due to the minimal number of discarded instances and the additional processing time for low-resolution WSIs.

\subsection{Further Analysis: the Impact of Dataset Size}
\label{subsec:further analysis}
Despite the performance gap between HDMIL$\dag$ and HDMIL on the Camelyon16 test set (depicted in~\cref{tab:comparative_results}), both methods show similar performance on the validation sets across all three datasets (shown in~\cref{tab:each_component} and~\cref{fig:mask_ratio}).
The drop in performance of HDMIL, on the Camelyon16 test set, is likely attributed to the bias introduced by selecting models for evaluation based on their validation performance. 
This bias can result in suboptimal performance on the test sets, especially when there is a distribution difference between the validation and test sets, which becomes more pronounced when dealing with smaller datasets.

\begin{table}[!h]
\setlength{\tabcolsep}{3pt}
\renewcommand{\arraystretch}{1.0}
\centering
\begin{tabular}{cc|cc|cc|cc}
\toprule
\multicolumn{4}{c|}{TCGA-NSCLC} & \multicolumn{4}{c}{TCGA-BRCA} \\
\hline
\multicolumn{2}{c|}{Valid Set} & \multicolumn{2}{c|}{Test Set} & \multicolumn{2}{c|}{Valid Set} & \multicolumn{2}{c}{Test Set} \\
\hline
25\% & 100\% & 25\% & 100\% & 25\% & 100\% & 25\% & 100\% \\
\hline
${95.84}$ & ${95.88}$ & ${95.46}$ & ${96.35}$ & ${94.55}$ & ${93.27}$ & ${90.06}$ & ${90.45}$ \\
\bottomrule
\end{tabular}
\caption{
The 10-fold average validation and test AUC scores of HDMIL, when the number of cases in the validation set is reduced to 25\% and the cases in the training and test sets are unchanged.
}
\label{tab:less_validation}
\end{table}

To demonstrate our point, we conducted specific experiments on TCGA-NSCLC and TCGA-BRCA, as shown in~\cref{tab:less_validation}.
It can found that when the number of cases in the validation set was reduced to 25\% of the original size, the performance of HDMIL on the validation set did not decrease significantly, while the models selected using these validation sets performed worse on the test sets. 
This finding demonstrates that the performance decline of HDMIL on Camelyon16 can be attributed to the small dataset size rather than inherent algorithmic flaws.

\section{Conclusion}
\label{sec:conclusion}
In this paper, HDMIL offers a novel approach for accelerating WSI classification while ensuring high classification accuracy.
By hierarchical-distillation, HDMIL efficiently filters out irrelevant patches within WSIs, significantly reducing inference time.
Extensive experiments demonstrate that HDMIL outperforms current state-of-the-art methods in both classification performance and inference speed.
To further improve MIL efficiency, we will explore strategies to alleviate the inference burden on the feature extractor and integrate them with HDMIL in the future.

{
    \small
    \bibliographystyle{ieeenat_fullname}
    \bibliography{main}
}



\end{document}